\documentclass[10pt,twocolumn,letterpaper]{article}

\usepackage{3dv}
\usepackage{times}
\usepackage{epsfig}
\usepackage{graphicx}
\usepackage{amsmath}
\usepackage{amssymb}
\usepackage{amsfonts}
\usepackage{subfigure} 
\usepackage{multirow}
\usepackage{multicol}
\usepackage[linesnumbered,algoruled,boxed,lined]{algorithm2e}
\usepackage{calrsfs}
\usepackage[pagebackref=true,breaklinks=true,letterpaper=true,colorlinks,bookmarks=false]{hyperref}
\usepackage[misc]{ifsym}

\threedvfinalcopy 

\def\threedvPaperID{} 

\begin{document}

\title{Real-Time 3D Shape of Micro-Details }

\author{Maryam Khanian$^1$(\rm{\Letter})\\
{\tt\small  khanimar@b-tu.de}
\and
Ali Sharifi Boroujerdi$^1$\\
{\tt\small shariali@b-tu.de}
\and
Michael Breu\ss$^1$\\
{\tt\small michael.breuss@b-tu.de} \\ \\
\footnotesize{$^1$Chair of Applied Mathematics, Brandenburg University of Technology (BTU), 03046 Cottbus, Germany}
}
\maketitle

\begin{abstract}
Motivated by the growing demand for interactive environments, we propose an accurate real-time 3D shape reconstruction technique. To provide a reliable 3D reconstruction which is still a challenging task when dealing with real-world applications, we integrate several components including \textit \emph{(i)} Photometric Stereo (PS), \textit  \emph{(ii)} perspective Cook-Torrance reflectance model that enables PS to deal with a broad range of possible real-world object reflections, \textit \emph{(iii)} realistic lightening situation, \textit \emph{(iv)} a Recurrent Optimization Network {\bf{(RON)}} and finally (v) heuristic Dijkstra Gaussian Mean Curvature {\bf{(DGMC)}} initialization approach. We demonstrate the potential benefits of our hybrid model by providing 3D shape with highly-detailed information from micro-prints for the first time. All real-world images are taken by a mobile phone camera under a simple set up as a consumer-level equipment. In addition, complementary synthetic experiments confirm the beneficial properties of our novel method and its superiority over the state-of-the-art approaches.
\end{abstract}
\section{Introduction}
3D reconstruction can be considered as one of the major areas for computer vision which provides not only 3D interpretation, but also the surface feature extraction. Photometric Stereo (PS) is a potential technique for this purpose. In PS, several input images taken from a fixed view point under different illumination directions are used for 3D reconstruction. 
The pioneers of the problem are Woodham \cite{woodham78} and Horn \textit et \textit al. \cite{HWS78}.
As it is shown by Woodham \cite{Woodham1980}, the orientation of a Lambertian surface can be uniquely determined when at least three input images are given. During last decades, PS has attracted wide attention in different practical domains such as industrial quality control \cite{Farooq2005,Tseng2013,Parkin2012,Kang2001}, face recognition \cite{{Yu2010},Zhou2007,Georghiades2001,Zafeiriou2001}, medical science \cite{{Parot2013},Tsumura2001}, texture classification \cite {Sathyabama2001,MLSmith1999,{MCGunnigle1999}} and estimation of weather conditions \cite{shen2009photometric}.
Another benefit of PS is to generate invariant features utilized in different kinds of recognition tasks \cite{Tang2012,{WLSmith2016}}.
\begin{figure}[t]
\centering
\includegraphics[width=0.48\textwidth]{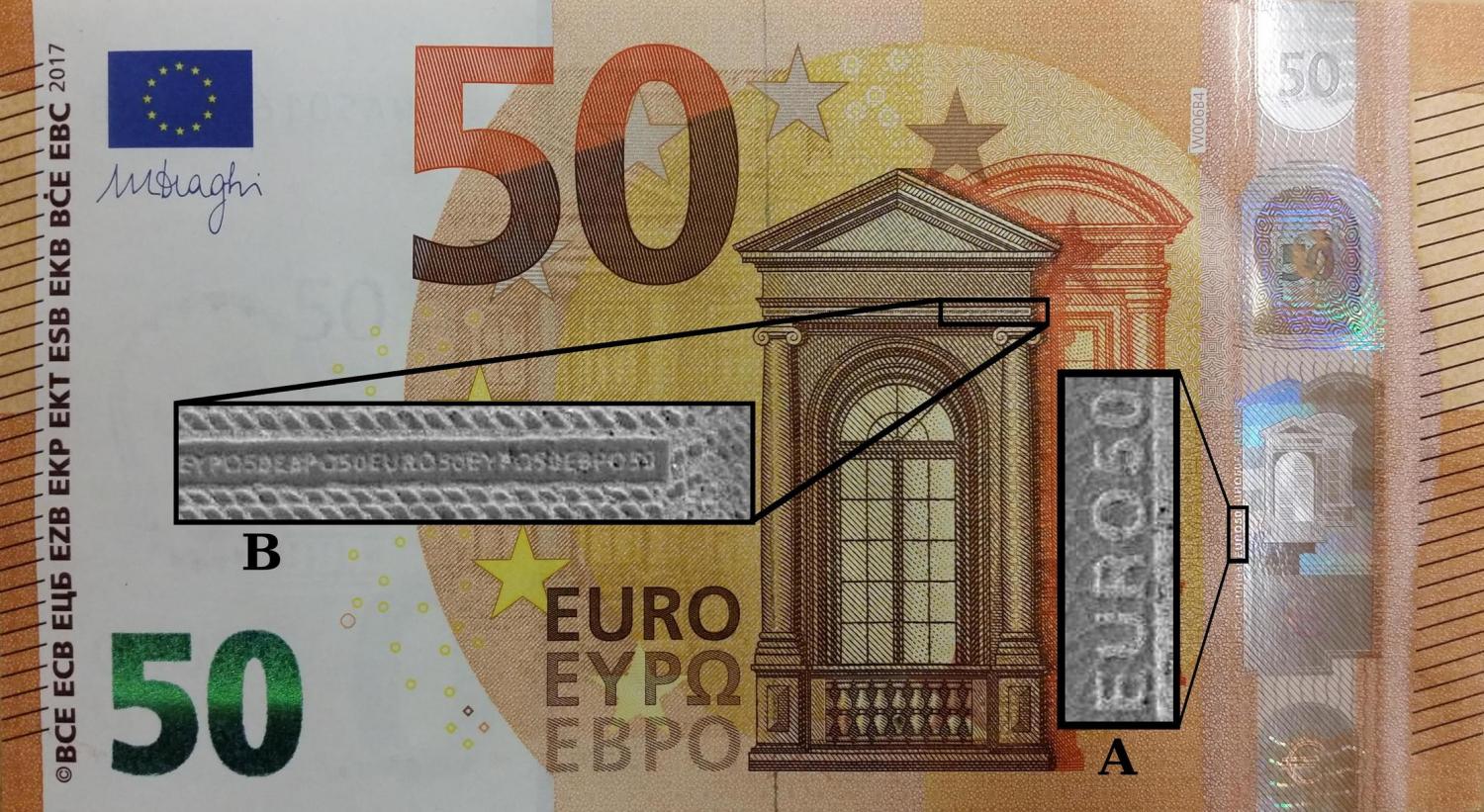}
\caption{A successful 3D reconstruction from very fine details of micro-prints on a 50 euro banknote shown in gray color rectangular parts of (A) and (B). As can be seen, these details are even invisible to human eyes, while our {\bf{real-time}} 3D reconstruction from these regions is able to reveal hidden information and features in high amount of details. This information recovery can be used in various applied areas such as detecting security items on financial documents for the fraud detection and also the quality control of any industrial productions that include delicate details such as printed circuits. 
\label{Euro}
}
\end{figure}
However, most of the later works on PS kept two basic building blocks of Woodham's approach, namely Lambertian surface reflection and orthographic camera projection.
For this reason, Lambert's law \cite{lambert1760} is commonly used in classic computer vision models, which implies that the surface reflects light equally in all directions and ignores the specular component. However, it may easily fails for surfaces whose reflection is non-Lambertian. Corresponding effects (e.g.\ specularities as occurring on shiny objects) are quite readily observable in real-world images and surfaces with not pure isotropic luminance. So, two major light reflection components including specular reflection as the high frequency variations and diffuse reflection as the low frequency component should be combined to provide a successful reflectance model \cite{{Plumbley2003},{Nayar1991},{Wollf1994}}. \\
In order to circumvent difficulties of specularities, commercial instruments for 3D reconstruction use sophisticated techniques such as white light interferometry or scanning focal microscopy, while these laboratory-based devices tend to be large, slow, or at the expense of \$100,000 or more.
In many works, non-Lambertian effects are considered as outliers.
In this context, Miyazaki \textit et \textit al.\ \cite{MHI10} suggested a median filtering to evade the influence of specular reflection and 
a preprocessing technique is presented by Yang \textit et \textit al.\ \cite{Yang2010} to remove specularities. A maximum-likelihood estimation \cite{Verbiest2008} and an expectation maximization \cite{Wu2010} are other approaches based on assuming specularities as the outliers. While Mukaigawa \textit et \textit al.\ \cite{MIS07} suggested a random sample consensus based approach in which only diffuse reflection is 
employed as the underlying model, Tang \textit et \textit al.\ \cite{TTW05} proposed a coupled Markov random field treating specularities as noise. An approach relying on a maximum feasible subsystem 
method is introduced by Yu \textit et \textit al.\ \cite{Yu2010}, where input images satisfying a Lambertian constraint are selected for reconstruction. In addition, a convex optimization technique is developed by Wu \textit et \textit al.\ 
\cite {WGS10} in which specularities are considered as deviations from the Lambertian assumption in the objective function. Zickler \textit et \textit al.\ \cite{IK12} provides an analyze to a subspace which is free from highlights for image transformation. 
Finally, Ikehata \textit et \textit al.\ \cite{IK12} suggested a regression procedure that deals with the specular component as a sparse error of an underlying purely diffuse reflectance equation. Mallick \textit et \textit al.\ \cite{{Mallick2005}} considered a transformation of the RGB color space to avoid specularity using the light source color which is not always available. An inpainting technique for the highlights correction is presented by Tan \textit et \textit al.\ \cite{Tan2003} using the color of illuminant and the diffuse component. In \cite{TanIkeuchi2003} and \cite {Shen2009} the pixel chromaticity direct analysis and specular-free image are used respectively to specularity correction in which the diffuse component is adjusted to a criterion. This criterion is an indicative of smooth transition of color between diffuse and specular regions and evaluated using a nonlinear shifting of specular pixels intensity and chromaticity. 
Johnson \textit et \textit al.\ \cite {Johnson2009} presented an elastomer using a sensor skin constructed
with a metal-flake pigment which is attached to the surface to change its BRDF.
However, in all these works, the shape information that is evidently contained in non-Lambertian effects 
(e.g.\ highlights) is discarded.
The orthographic projection as the second mentioned ingredient is realistic when objects are far away from the camera, but not if they are close since then perspective effects grow to be important. In addition, when a surface is projected orthographically, it can be reconstructed only up to a generalized bas-relief transformation. However, the perspective projection as proved in \cite{papa}, can remove the bas-relief ambiguity and provide the uniqueness of the reconstructed surface.
There is only a small number of previous papers dealing with perspective models in PS. 
To our knowledge, the first of such models has been developed by Lee and Kuo \cite{Lee1996}.
They approximated the depth function as a linear combination of individual element basis functions.
Other perspective methods are applied by Galo and Tozzi \cite{GT96}, Tankus and Kiryati \cite{TK05} and
Mecca \textit et \textit al. \cite{RMECCA15}, while the Lambertian reflectance is applied in all these researches.
Concerning more general non Lambertian reflectance models based on perspective PS, Khanian et al. proposed the application of perspective Blinn-Phong model \cite{Blinn77, Phong75} for the signal reconstruction and corresponding lightening conditions in \cite{khanian} and for the 3D reconstruction with different perspective techniques in \cite{Khanian2017}. An empirical approach is also considered in \cite{MeccaRoCr2015, meccayvain}. However, for the derivation of the final applied model, purely diffuse and purely specular reflections should be decoupled in \cite{MeccaRoCr2015, meccayvain} . Moreover, the proposed approach in \cite{MeccaRoCr2015} relies on the Fast Marching method \cite{FM} where the unknown depth value of the center point should be provided in advance which makes difficulty for real-world applications. Some state-of-the-art in PS methods in which the information of depth should be provided in advance, can be pointed as \cite{{Chandraker2013},{Yvain2015}} which require the depth to be known on the boundary, and \cite{meccayvain} that needs the mean value of depth. In contrast, we do not need any information of depth to be given in advance in our approach as also proved in Figure \ref{Euro}.
\subsection{Our contributions}

\begin{figure}[t]
\centering
\includegraphics[width=0.35\textwidth]{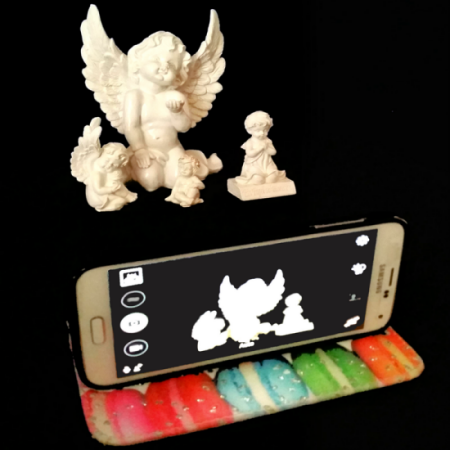}
\caption{A complex real scene including several objects with different sizes captured by a mobile phone with a simple setup which is used for our 3D shape reconstructions.
\label{fig1}
}
\end{figure}

In this paper, we provide the following innovations:\\
1. We proposed an accurate hybrid {\bf{{real-time}}} 3D reconstruction approach which merges the benefits of perspective projection and non-Lambertian reflectance model motivated by Cook-Torrance reflectance \cite{Cook82} to handle complex reflecting characteristics. To the best of our knowledge, this is the first work for the real-time PS which also regarded the application of perspective Cook-Torrance model.\\
2. Introducing a {\bf{Recurrent Optimization Network (RON)}} which provides a robust PS framework to obtain a highly-detailed 3D shape from even {\bf{micro-print}} texts where their visibility is hard for the human visual system as can be seen in Figure \ref{Euro}. In this Figure, we could reconstruct 3D shape of two regions of {\bf{A}} and {\bf{B}} with high details, while these regions are very tiny microtexts. The information hidden in these areas became completely clear by our approach. As a result, our technique can be used in different applied areas.\\ 
3. Since the performance of many optimization techniques is highly-dependent on the initialization process, we provided a {\bf{Dijkstra Gaussian Mean Curvature (DGMC)}} technique to find anchor points which can be applied as a key point to offer more effective optimization approach.\\        
4. Furthermore, we extended our proposed approach to deal with realistic lightening situation using spatially dependent lighting.\\
5. We show that our approach is able to provide the real-time 3D shape of objects as captured by an ordinary mobile phone camera without employing controlled laboratory conditions (cf.\ Figure \ref{fig1}). So, our PS process is conducted by the consumer-level equipment where specialized equipment and professional laboratory settings are not required. Our instruments are easily available for general consumers as should be expected for an ideal solution. Consequently, our method can be readily employed in real-world situations. We demonstrate that this paper provides the basis for a potentially useful 3D reconstruction technique. This improvement breaks inhibitor limitations of a controlled setup and the necessity of working with specific scenes (e.g. Lambertian materials) and turns mobile phone into powerful 3D shape reconstruction tools.\\
6. Finally, our approach also provides non-uniform colorful albedo from images with diverse color intensities as shown in Figure \ref{albedo}.\\ 
All these attempts greatly advance the applicability of photometric stereo to various applications,
especially for reconstructing 3D surfaces from very tiny structures like micro-prints.

\vspace{-1.5ex} 
\section{Cook-Torrance reflectance model}
The Cook-Torrance reflectance as a physically based model designed to account for the spectral composition of specularity and cope with the problem of modeling reflections of realistic materials \cite{Cook82}. This physically based model agrees well with actual measured BRDFs \cite{kurt}. Like recent popular theoretical models, this reflectance model considers this assumption that the large scale BRDF is a collection of microfacets as shiny V-grooves with random sizes and orientations \cite{kurt}. This hypothesis leads to the microfacet theory for non-perfectly-smooth surfaces. These surfaces are essentially height fields, where the distribution of facet is described statistically. So, the main component of these models is an expression for the distribution of microfacet normals. This property enables them to effectively model many real-world surfaces \cite{{PBRT}} in which, mirror-like facets are modeled by the Cook-Torrance reflectance to retrieve the local shape from specularity. 
The Cook-Torrance image irradiance equation is defined as follows:\\
\begin{equation}
\hspace{-38mm}I(x,y)=\underbrace{k_d \frac{L\cdot N(x,y)}{\left\| L \right\|\left\|{N(x,y)}\right\|}l_d}_{diffuse \; term} \nonumber\\
\end{equation} 
\begin{eqnarray}
+&\underbrace{ \frac{G(x,y,z) D(x,y,z) F(x,y,z)}
{4\frac{V(x,y,z)\cdot N(x,y)}{\left\|{V(x,y,z)}\right\|\left\|{N(x,y)}\right\|}\frac{L\cdot N(x,y)}{\left\|{L}\right\|\left\|{N(x,y)}\right\|}}}_{specular \; term}
\label{eq-cook1}
\end{eqnarray}
where
\begin{eqnarray}
\hspace{-10mm}G(x,y,z)=
\min
\Bigg\{1,T(x,y,z), R(x,y,z)
\Bigg\},
\label{eq-cook2}
\end{eqnarray}
\begin{eqnarray}
\hspace{-3mm}T(x,y,z)=\frac{
\frac{2 H(x,y,z)\cdot N(x,y)}{\left\|{H(x,y,z)}\right\|\left\|{N(x,y)}\right\|}
\frac{V(x,y,z)\cdot N(x,y)}{\left\|{V(x,y,z)}\right\|{\left\|{ N(x,y)}\right\|}}}{Q(x,y,z)},
\label{eq-cook3}
\end{eqnarray}
\begin{eqnarray}
\hspace{-13mm}R(x,y,z)=\frac{\frac{2 H(x,y,z)\cdot N(x,y)}{\left\|{ H(x,y,z)}\right\|\left\|{ N(x,y)}\right\|}
\frac{L\cdot N(x,y)}{\left\|{ L}\right\|{\left\|{ N(x,y)}\right\|}}}{Q(x,y,z)},
\label{eq-cook3B}
\end{eqnarray}
\begin{eqnarray}
\hspace{-20mm}Q(x,y,z)={\frac{V(x,y,z)\cdot H(x,y,z)}{\left\|{ V(x,y,z)}\right\| \left\|{ H(x,y,z)}\right\|}},
\label{eq-cook4}
\end{eqnarray}
\begin{eqnarray}
\hspace{-12mm}D(x,y,z)=\frac{1}{\pi{m^2}{\cos^4(\alpha)}}
\exp\left(
\frac{-\tan^2(\alpha)}{ m^2}
\right),
\label{eq-cook5}
\end{eqnarray}
\begin{eqnarray}
\hspace{-21mm}\alpha=\arccos\left(\frac{H(x,y,z)\cdot N(x,y)}{\left\|{ H(x,y,z)}\right\|\left\|{ N(x,y)}\right\|}\right),
\label{eq-cook6}
\end{eqnarray}
\\
and
\vspace{10mm}
\begin{equation}
 \begin{aligned}
\hspace{-44mm}F(x,y,z)= f_{\lambda}\nonumber
\end{aligned}
\end{equation}
\begin{equation}
 \begin{aligned}
+(1-f_{\lambda})\left( 1-
\frac{H(x,y,z)\cdot V(x,y,z)}{\left\|{H(x,y,z)}\right\|\left\|{ V(x,y,z)}\right\|}\right)^5
\label{eq-cook7}
\end{aligned}
\end{equation}
\noindent In this notation, $I(x,y)$ and $ N(x,y)$ are the intensity and surface normal at pixel $(x,y)$, respectively.
The parameter $k_d$ is the diffuse material parameter (albedo) and $m$ is the root mean square slope of the surface microfacets (i.e.\ the surface roughness).\\
It is worth to mention that in expression (8), $H$ is used instead of $N$ as suggested in \cite{{Hoffma2013}}.
Let us emphasize that as opposed to an orthogonal viewer direction $(0, 0, 1)^\top$ 
which is commonly used, we model here a pinhole camera. 
Therefore, the viewing vector $V$ depends explicitly on $(x,y)$ 
and also on the depth $z$.
This model is powerful enough to capture the geometrical properties of the scene as recorded by the camera. Moreover, its images resemble what would be seen by human visual system. Hence, pinhole camera is also referred to as perspective camera model in graphics and vision literatures \cite{Ye2013}.\\
The vector $H$ is defined to be the halfway vector between the viewing vector $V$ and the light vector $L$, as the direction to which the microfacet normals need to be oriented
to reflect $L$ into $V$. The light source intensity is denoted by  $l_d$. $F$ is the Fresnel term including $f_\lambda$ which is the reflection coefficient (the reflectance of the surface at normal incidence).
Here, the distribution of facets is described by $D$. 
It should be noticed that in PS, it is necessary to apply a set of non-coplanar light vectors $L_i$ (corresponding to the input images $I_i$) as also investigated in \cite{Woodham1980,horn89,horn86,davis}. 

It is worthwhile noting that in all our experiments (including synthetic and real-world scenes and even for specular images), we will still follow Woodham \cite{woodham78} 
and use only {\bf{three input images}} $I_{i}, i = 1, 2, 3$. We consider it an indication of the 
robustness and potential practical usefulness of our method that it
works effectively with such a low number of input images.
Considering (1) for multiple images leads to a highly non-
linear system of equations for the unknown surface normal
vectors that should be solved.
\section{Perspective projection and intrinsic parameters of pinhole camera model} 
 
Under the perspective projection model, the surface parametrization is performed as follows:

\begin{equation}
S(x,y) = \begin{bmatrix}
       { \frac{-xz(x,y)}{f}} &\\
        {\frac{-yz(x,y)}{f}}\\
         {z(x,y)}           
     
     \end{bmatrix}
\label{eq-pers}
\end{equation}
The normal vector as the cross product of the partial derivatives of the surface is computed as:

\begin{equation}
N(x,y) = \begin{bmatrix}
       {{\frac{z(x,y)}{f}}\nabla  z(x,y)} &          \\[0.3em]
      {\frac{z(x,y)}{f^2}(\nabla  z(x,y).(x,y)+z(x,y))} & \\[0.3em]
     \end{bmatrix}
\label{eq-pers}
\end{equation}
where $ f$ is the focal length and $\nabla  z(x,y)=(z_x,z_y) $ is the gradient field. On the other hand, to implement the perspective projection of the pinhole camera model, the intrinsic camera parameters should be involved which leads to the CCD camera \cite{hart}. To this aim, we apply the following transformation to convert the homogeneous representations pixel coordinate $ X=(x,y,1)$ to image coordinate $\chi=(\mu,\nu, f)$:

\begin{equation}
X = \Gamma (\frac{1}{f}) \chi,
\end{equation}
\noindent
where $\Gamma$ is the matrix of intrinsic parameters of the camera expressed as:
\begin{equation} 
\Gamma = \begin{bmatrix}
       \psi_x & \xi&  \delta_x \\
      0 & \psi_y &\delta_y \\
       0 & 0 &1 
     \end{bmatrix}
\label{eq-pers}
\end{equation}
\noindent
Here, $\psi_x$ and $\psi_y$ are the focal length in x and y directions.
$ \xi$ describes the skew effect and $(\delta_x,\delta_y)$ denotes the principal point or focal point. We neglect the skew parameter ($\xi$ = 0) as it is zero for most of normal cameras \cite{hart}.

\section{Lightening sensitivity analysis} 

Different types of light sources can be defined based on the energy emission from a single point in space with some angularly varying distribution of outgoing lights. At first, we considered the distant light, also known as the directional light or sparse light. This source of energy is an emitter which deposits illumination from the same direction at every point in space. It can be considered also as a light at infinity (e.g. sun). Furthermore, in this kind of light sources, energy strength does not depend on the distance between the light source and the object \cite{{PBRT}}.
In addition, we also equipped our perspective Cook-Torrance model with spatially varying lightening in which the point-wise light source implies how the light varies from one point to another. This assumption facilitates the independent control of intensity. To this aim, a three dimensional vector field of the light $L$ should be imposed instead of a directional light vector. In this case, the rate at which the light falls off w.r.t. the distance should be induced to make the model more consistent with the actual physical aspects. Integrating all the mentioned points, the new spatially dependent light vector $L$ is defined as:
\begin{eqnarray}
L(x,y)= \Pi
 {\frac{  \Upsilon - S(x,y)}{ \left\|{ \Upsilon - S(x,y)}\right\|}}
\label{eq-cook4}
\end{eqnarray}
with the light attenuation $\Pi $:
\begin{eqnarray}
\Pi = 
 {\frac{l_d}{ {\left\|{ \Upsilon - S(x,y)}\right\|}^{2}}}
\label{eq-cook4}
\end{eqnarray}
where $ \Upsilon $ is the light source position.
\noindent
Since we developed the perspective projection, the surface parametrization as in (9) is imposed in $ S(x,y)$ for spatially point light source. As a result, the light direction depends not only on the point (x,y), but also on the depth of the surface itself. So, in this case, an initial guess for the depth of the surface is required. 
\section{Optimization process}
Using equation (1) for three input images (as the minimum number of input images in PS), we constitute a system of equation $F(X)=0$, where $F:{\mathbb{R}}^n \longrightarrow {\mathbb{R}}^m$ is defined as $F(X)= [f_h], $ $h=1,2,3$ and formed by equations given by (1) corresponding to each input image $I_h(x,y),$ $h=1,2,3$ as: 
\begin{equation}
f_{h}=I_h(x,y)-(diffuse\hspace{1mm}term) - (specular\hspace{1mm}term) = 0
\end{equation} 
To estimate the normal field, we should solve the optimization problem as follows:\\ 
 \phantom{x}\hspace{3ex}$X^{*}=arg min\hspace{3mm} \frac{1}{2} {\|F(X)\|}^{2}=\frac{1}{2}$$\sum_{h=1}^{3} f_h^{2}$\\
 In optimization process, we update the solution using an iterative procedure:
 \begin{equation}
X^{k+1}:=X^{k}+{d}^k,
\
\label{eq-opti}
\end{equation}
Here, $d^{k}$ is the search direction in iteration $k$. The main difference between various optimization approaches is caused by the definition of $d^{k}$.
\subsection{Quasi-Newton with BFGS updating}
In this section, we investigate various procedures that we applied for the nonlinear system of equations described by $F(x)=0$. Among many iterative methods, Newton's method is one of the popular approaches. But this method has some drawbacks which can be removed by its modified versions. Here, we used the updated formula known as Quasi-Newton with BFGS suggested independently by \cite{Broyden1970,Fletcher1970,Goldfarb1970,Shanno1970}, which as explained in \cite{{Powell1970}}, has much better performance than other modified versions of Newton's method. In this approach, the Quasi-Newton condition (secant condition) is used to find the search direction $d^{k}$:
\begin{equation}
{d}^k:=-(B^{-1})^{k}{J{(X}^{k})}
\label{eq-quasi-1}
\end{equation}
where $B$ and $J(X)$ are the Hessian matrix approximation and the Jacobian matrix of $F(X)$, respectively. The following expression is known as Quasi-Newton condition or the BFGS formula for updating $B$:
\begin{equation}
B^{k+1}=B(X^{k})+\frac{Y^{k}(Y^{k})^T}{({\Theta}^{k})^TY^{k}}-\frac{U^{k}(U^{k})^T}{(\Theta^{k})^TU^{k}}
\
\label{eq-BFGS}
\end{equation}
with
\begin{equation}
 Y^{k}=J{(X^{k+1})}-J{(X^{k})},\\
  \end{equation}
   \begin{equation}
\hspace{-10mm} {\Theta}^{k}=X^{k+1}-X^{k},\\
  \end{equation}
   \begin{equation}
  \hspace{-21mm}U^{k}=B^{k}{\Theta}^{k}
 \end{equation}
\subsection{Levenberg-Marquardt}
Another strategy that we considered is the Levenberg-Marquardt method \cite{{leven44},{Marq63}} with the search direction determined as:
\begin{equation}
d_{k}=-(J(X^{k})J(X^k)^{T}+ \lambda_k I)^{-1} J(X^{k})F(X^{k})
\label{eq-levenberg}
\end{equation}
where $ \lambda _k> 0$ and the $ (J(X^{k})J(X^k)^{T}+ \lambda_k I)^{-1}$ is positive definite.

\subsection{Powell's Dog Leg}
In this technique, a strategy is proposed to choose optimization steps based on a parameter called radius of trust region $\Delta$. Three search directions are applied as follows:
\begin{equation}
\hspace{-36mm}{h_{gn}}^k:=-J(X^{k})^{-1}F(X^{k}),\\
\label{eq-powell-1}
\end{equation}
\begin{equation}
\hspace{-36mm}{h_{sd}}^k:=-\alpha J(X^{k})^{T}F(X^{k}),
\label{eq-powell-2}
\end{equation}
\begin{equation}
\hspace{-14mm}{h_{dl}}^k:={h_{gn}}^k,\hspace{21mm} if {\left\|{h_{gn}}^k\right\|}\leq {\Delta}
\label{eq-powell-3}
\end{equation}
\begin{equation}
\hspace{-12mm}{h_{dl}}^k:=\frac{\Delta}{\left\|{h_{sd}}^k\right\|}{h_{sd}}^k,\hspace{10mm} if {\left\|{\alpha} {h_{sd}}^k\right\|} \geq \Delta
\label{eq-powell-4}
\end{equation}
\begin{equation}
\hspace{-11mm}{h_{dl}}^k:=\alpha{h_{sd}}^k+\beta({h_{gn}}^{k}-\alpha{{h}_{sd}}^{k}),\hspace{8mm} o.w.
\label{eq-powell-5}
\end{equation}
where
\begin{equation}
\alpha:=\frac{\left\|J(X^{k})^{T}F(X^{k})\right\|^{2}}{\left\|J(X^{k})J(X^{k})^{T}F(X^{k})\right\|^{2}}
\label{eq-powell-6}
\end{equation}
Here, we have also used the strategy applied in \cite{{Madsen1970}} to choose $\beta$ and $\Delta$. The general process of mentioned algorithms is illustrated in the last section.
\subsection{Dijkstra Gaussian Mean Curvature (DGMC) technique}
As a heuristic search to find a proper initial value for the normal field, we have devised a technique that consists of two phases. At the first stage, we employ the Dijkstra algorithm \cite{dij} to find the nearest point in the outer highlight boundary $S$ to each specular pixel $p$ . This point is denoted as $q$.\\
\noindent
In the second step, we compute the \textit{Gaussian Curvature (GC)} and the \textit{Mean Curvature (MC)} properties of $p$ and $q$, separately:
\begin{equation}
GC = K_1 \times K_2, \quad MC = \frac{K_1+K_2}{2}
\end{equation}
where $K_1$ is the smallest and $K_2$ is the biggest eigenvalues of the Hessian matrix of the pixel local neighboring area.
\noindent
Having those parameter for both points allows us to define their local geometric similarity conditions as follows:
\begin{equation}
GC_q = GC_p \pm 5 \%
\end{equation}
\begin{equation}
MC_q = MC_p \pm 5 \%
\end{equation}
In case that $q$ does not meet similarity conditions, the next nearest member of $S$ to $p$ will be nominated.
\subsection{Recurrent Optimization Network (RON)}
We will introduce an intermittent optimization strategy which allows updating not only the surface normal but also the albedo and roughness parameter. At first, we consider a constant albedo and also a constant roughness so that we can obtain the surface normal. After obtaining the normal field, we update the albedo values. Furthermore, the roughness parameter is updated as well. To this aim, we use the following formulation proposed in \cite{gullon} to update the roughness parameter $m$, where we already obtained the depth map $z$ by integrating the normal field.
\begin{equation}
m = \sqrt{\frac{1}{n}\sum_{x=1}^{n} {\sum_{y=1}^{n}}\Big(z(x,y)-\bar{z}{(x,y)}\Big)}
\end{equation}
Here, $\bar{z}{(x,y)}$ is the mean surface depth, and $n$ is the number of pixels. The whole algorithm of mentioned RON procedure is illustrated in Algorithm 1.\\ Thanks to this concise optimization network with back-tracing steps and DGMC technique which are integrated into the photometric stereo algorithm, our method enables PS to work in diverse applications by removing restrictions such as diffuse objects, orthographic projection, laboratory set up and professional equipment.
\begin{algorithm*}[t]
\LinesNotNumbered
\SetKwData{Left}{left}
\SetKwData{This}{this}
\SetKwData{Up}{up}
\SetKwFunction{Union}{Union}
\SetKwFunction{FindCompress}{FindCompress}
\SetKwInOut{Inputf}{Primary input}
\SetKwInOut{Inputs}{Secondary inputs}
\SetKwInOut{Output}{output}
\Output{Normal vectors, colorful albedo, roughness value and the depth map}

\BlankLine
\While{{stopping criterion is not satisfied }}{
	{
	\begin{enumerate}
	\vspace{2mm}
		\item Compute normal vectors from eq (1) using the proposed DGMC technique and one of Algoritm 2,  Algoritm 3 or Algoritm 4 by the constant albedo and roughness\\
		\item Update albedo from eq (1) by the obtained normal field using the same choice of Algoritm 2,  Algoritm 3 or Algoritm 4\\
		\item Integrate the normal field using the proposed approach in section 5.6. to obtain depth map $z$ \\
		\item Update {the roughness value by applying }
		equation (32)\\
		\item Update the normal field from eq (1) using the recent albedo and roughness 
	\end{enumerate}  					
	}	
}{}

\caption{Recurrent optimization network }\label{algo_disjdecomp}
\end{algorithm*}

\subsection{Integration}
As the integrator of the normal field, the energy minimization of the following functional should be considered: 
\begin{equation}
\mathcal{F}(u) = \iint_\Omega{\|\nabla z(x,y)-N(x,y)\|}{^2} dx\;dy
\end{equation}
which leads to the Euiler-Lagrange equation $\Delta z= \nabla.N$.
\\
In order to solve the mentioned minimization problem, we applied Generalized Minimal RESidual (GMRES) with the initial solution of Simchony \cite{sim}.

\section{Experiments} 
In the case of synthetic experiments, to perform a quantitative evaluation, we will make use of Mean Angular Error of Normal vectors (MAEN) in degrees and also Mean Squared Error (MSE) of depth.
\subsection{\bf\textit {Tests of accuracy on synthetic data}}
\begin{figure*}[ht]
\centering{
\begin{tabular}{cc}
\vspace{3mm}
\includegraphics[width=0.38\textwidth]{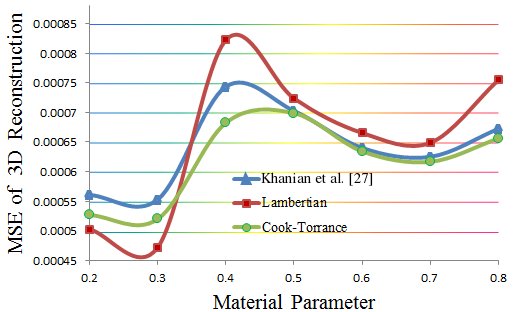}&
\includegraphics[width=0.38\textwidth]{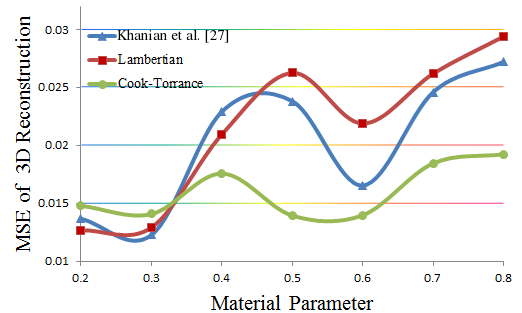}\\ 
\vspace{3mm}(a) Soldier & \vspace{3mm} (b) Horseman\\
\includegraphics[width=0.38\textwidth]{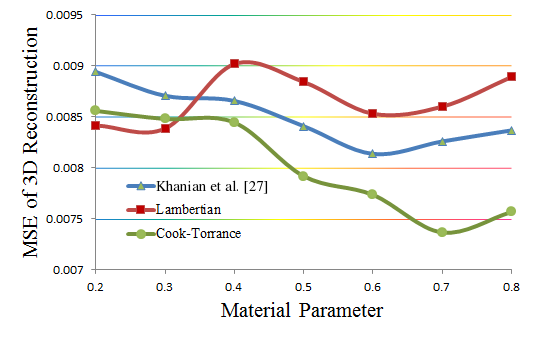}&
\includegraphics[width=0.38\textwidth]{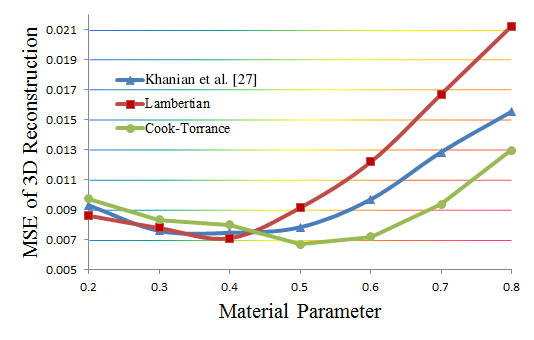}\\ 
(c) Fighter & (d) Coping\\
\end{tabular}
}
\vspace{3mm}
\caption{ Comparison with \cite{Khanian2017} and results obtained by Lambertian reflectance (as the most common model applied in PS) in presence of specularity. For a fair evaluation, these experiments are performed against various material parameter ($k_s$) for different input images.  
\label{fig2}
}
\vspace{-2mm}
\end{figure*}

In the following, we discuss our results obtained for the synthetic test images as displayed in Figures \ref{fig4}, \ref{fig5} and \ref{fig6}. 
All images are rendered using Blender software. The 3D model of Oldman and Woman images are obtained from \cite{Rishoo} and \cite{Woman} respectively. 3D models of the other synthetic images are publicly available at \cite{yobi}. All images except Coping 
($256\times 200$) are in the size of $256 \times 256$. All implementations are performed on an Intel Core i7 processor with 8 GB of RAM.
\\ \\
\textbf{Comparison with other BRDFs:}\\
In Figure \ref{fig2} documenting the first set of experiments, we represent
the results of comparison between the recent approach suggested by Khanian et al. \cite{Khanian2017} and also the results of Lambertian reflectance model (as the most common model applied in PS). The model presented in \cite{Khanian2017} is based on the Blinn-Phong reflectance.
We extended our experiments to a wide range of the specular material parameter $k_s = 1-k_d$ to evaluate how the varying $k_s$ influence the error of 3D reconstruction. It can be noticed that the output of our method achieves higher rate of accuracy over \cite{Khanian2017} and Lambertian model as $k_s$ increases. Whereas, there are some fluctuations when $k_d \gg k_s$ or $k_d \ll k_s$. As illustrated, error rates of presented models are almost similar for low values of $k_s$. However, as $k_s$ increases, the superiority of our model over Lambertian and \cite{Khanian2017} is significant. This observation agrees well with our motivation for focusing on the high frequency variations and model these components explicitly.
One particular interpretation for the sharp ascending trend of the error rate for \textit{Coping} input image (in the case of higher values of $k_s$ parameter) is the presence of the flat background region which leads to high amount of shiny highlighted areas in input images. As for the difference between \cite{Khanian2017} and Lambertian model, we find that the Lambertian still performs a little better on very low values of $k_s$; however, when $k_s$ is larger than 0.3, \cite{Khanian2017} improves accuracy much more than Lambertian model. 
As indicated, the MSE comparing our 3D reconstruction and
ground truth data, demonstrates that our method is capable of producing accurate 3D reconstructions for the 
indicated images with specularity.
The qualitative evaluations can be also seen as recovered surfaces shown in Figure\ \ref{fig4}. Our faithful 3D reconstructions with a high amount of details even for tiny surfaces containing low-depth details (e.g. Coping) represent the proficiency of our approach.\\ \\
\textbf{Optimization procedures:} \\
As the next experiment shown in Figure \ref{fig3}, we compared various optimization procedures (applied in first and second steps of our RON technique) for several surfaces with diverse $k_s$. It turns out that Dog Leg method achieves higher accuracy in all cases. Although, BFGS and Levenberg approaches follow the same trend for lower values of $k_s$ (Levenberg method can slightly outperform BFGS technique for low values of $k_s$) BFGS performs better as $k_s$ increases.\\ \\
\noindent{\bf{Lightening analysis:}}\\
To investigate the 3D reconstruction sensitivity w.r.t. the lightening model, we applied different PS frameworks. Quantitative and qualitative results of this investigation are shown in Table \ref{Table II} and Figure \ref{fig5}, respectively. To this end, we evaluated our approach regarding different available lightening conditions.
In the first experiment, we consider the distant light source leads to perform directional-light-based PS referred as DLPS. The output images of this simulation resemble the real-world images that are illuminated by sun. The second evaluation is conducted to simulate spatially dependent light resulting in point-wise-light-based PS denoted as PLPS. 
We observed that the higher accuracy is achieved by applying the point-wise light source and adjusting lightings based on spatially dependent light as presented in (13) and (14). Figure \ref{fig5} reveals that DLPS has problem to recover some areas since capturing images by directional-light leads to more shadows as shown in Figure \ref{fig5}. On the other hand, using the point-wise light source in the scene and formulating PLPS can provide the higher amount of accuracy (cf. Table \ref{Table II}).
\begin{table*}[t]
\centering
\begin{tabular}{c|c|c|c|}
\multicolumn{4}{c}{\hspace{15mm}Coping}\\\hline
 Methods & MAEN     &     MSED & Time  \\ 
\hline
\hline 
      
\multicolumn{1}{ |c| }{\bf{Our}}  & 0.4163& 0.0070 & 0.0485    \\ 

\multicolumn{1}{ |c| }{\cite{IK14}}  & 22.5992 & 0.1879 & 40.9955     \\ 

\multicolumn{1}{ |c| }{\cite{Barsky2003}}  & 12.9628 & 0.0424 & 58.4178     \\ 

\multicolumn{1}{ |c| }{\cite{MeccaRoCr2015}}  & 15.2238 & 0.0495 & 1.3645    \\ 

\multicolumn{1}{ |c| }{\cite{Woodham1980}}  & 11.9886 & 0.0409 & 1.4385     \\
\hline
&\multicolumn{3}{c}{Horseman}\\\hline
 Methods & MAEN     &     MSED & Time  \\
\hline
\hline
\multicolumn{1}{ |c| }{\bf{Our}}  & 0.7241& 0.0138 & 0.0470     \\ 

\multicolumn{1}{ |c| }{\cite{IK14}}  & 24.1228 & 0.0511 & 37.5942     \\ 

\multicolumn{1}{ |c| }{\cite{Barsky2003}}  & 14.2859 & 0.0306 & 52.3987     \\ 

\multicolumn{1}{ |c| }{\cite{MeccaRoCr2015}}  & 10.3619 & 0.0356 & 1.2781     \\ 

\multicolumn{1}{ |c| }{\cite{Woodham1980}}  & 4.2274 & 0.0424 & 1.3470   \\

\cline{1-4}

\end{tabular}
\begin{tabular}{|c|c|c|}
\multicolumn{3}{c}{Soldier}\\\hline
   MAEN     &     MSED & Time   \\ 
\hline
\hline 
      
0.2045 &   0.0007 & 0.0320    \\ 

29.6263 &  0.0455 & 32.1078     \\ 

19.8908 & 0.0205 & 45.9023     \\ 

7.2661 & 0.0131 & 1.0775    \\ 

5.2925 & 0.0113 & 1.1977    \\
\hline
\multicolumn{3}{c}{Fighter}\\\hline
  MAEN     &     MSED & Time   \\
\hline
\hline
0.5948 & 0.0079 &0.0460   \\ 

15.4457 & 0.0279 & 28.2307     \\ 

10.7417 & 0.0107 & 39.9016     \\ 

4.8770 & 0.0133 & 0.9929    \\ 

4.8039 & 0.0114 & 1.0205     \\ 

\cline{1-3}
\end{tabular}
\begin{tabular}{|c|c|c|}
\multicolumn{3}{c}{Militaryman}\\\hline
   MAEN     &     MSED & Time   \\ 
\hline
\hline 
      
0.2659 & 0.0053  & 0.0429    \\ 

24.3848 & 0.06356 & 40.4892    \\ 

19.9172 & 0.0430& 57.1407    \\ 

3.9674 & 0.0399 & 1.2868     \\ 

5.4562 & 0.0409 & 1.4496     \\
\hline
\multicolumn{3}{c}{Carpenter}\\\hline
  MAEN     &     MSED & Time   \\
\hline
\hline
0.2761 & 0.00308 & 0.0420     \\ 

23.2172 & 0.03018 & 39.9192     \\ 

21.1159& 0.02331 &29.1174     \\ 

4.8680 & 0.01192 & 0.9669     \\ 

1.0557 & 0.00915& 1.4744    \\ 

\cline{1-3}
\end{tabular}
\begin{tabular}{c|c|c|c|}
\multicolumn{4}{c}{\hspace{16mm}Oldman}\\\hline
 Methods & MAEN     &  MSED   & Time   \\ 
\hline
\hline 
      
\multicolumn{1}{ |c| }{\bf{Our}}  & 0.6734 & 0.0124 & 0.0466     \\ 

\multicolumn{1}{ |c| }{\cite{IK14}}  & 18.5054 & 0.0779 & 37.3778     \\ 

\multicolumn{1}{ |c| }{\cite{Barsky2003}}  & 13.2650 & 0.0400 & 51.4226     \\ 

\multicolumn{1}{ |c| }{\cite{MeccaRoCr2015}}  & 5.1195 & 0.0290 & 1.1693     \\ 

\multicolumn{1}{ |c| }{\cite{Woodham1980}}  & 3.9751 & 0.0222 & 1.3908    \\
\cline{1-4}

\end{tabular}
\begin{tabular}{|c|c|c|}
\multicolumn{3}{c}{woman}\\\hline
   MAEN     &     MSED & {\cite{MeccaRoCr2015}}   \\ 
\hline
\hline      
0.6157 & 0.0128 & 0.0393    \\ 

12.0515 &  0.11641 & 21.0516     \\ 

8.5208 & 0.0538 & 28.3929     \\ 

3.3653 & 0.04354 & 1.0492    \\ 

2.6305 & 0.0350 & 0.7499    \\
\hline
\end{tabular}
\vspace{2.0ex}
\centering 
\caption{Comparison of several PS methods w.r.t Mean Square Error of Depth (MSED), Mean Angular Error of Normal vectors (MAEN in degrees), and Time in seconds. Innovations aggregated in our approach leads to a significant improvement regarding obtained normal vectors and depth as well as computational Time.}
\label{Table I}
\end{table*}
\begin{figure*}[ht]
\begin{tabular}{cc}
\centering
\vspace{3mm}
\includegraphics[width=0.38\textwidth]{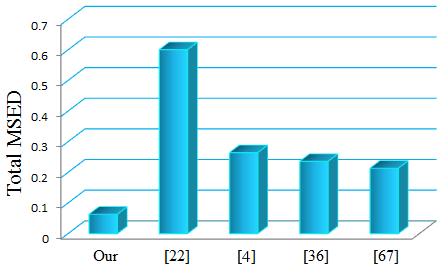}&
\hspace{22mm}\includegraphics[width=0.38\textwidth]{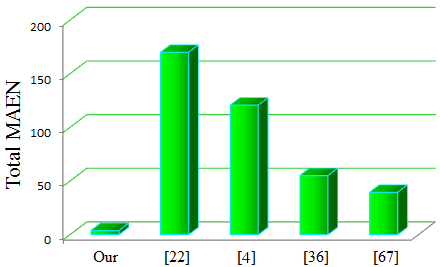}\\ 
\end{tabular}
\vspace{3mm}
\caption{Left: Total MSED  and Right: Total MAEN gained by our approach and compared methods on all images mentioned in Table 1. As it can be seen in the graph, the
proposed method can be also favored due to its higher accuracy w.r.t both obtained normal vectors and depth map. To illustrate, our technique can provide accuracy improvement in normal field extraction over {\cite{IK14}} by 97.78 \%  and over {\cite{Barsky2003}} by 96.87 \% . Furthermore, this improvement in accuracy of depth reconstruction by our scheme over {\cite{IK14}} is 89.51 \% and over {\cite{Barsky2003}} is 76.17 \%.   
\label{diagrams}
}
\vspace{-2mm}
\end{figure*}
\begin{table}[t!]
\begin{tabular}{ |c|c|c|c| } 
\hline
{\cite{IK14}} & {\cite{Barsky2003}} & {\cite{MeccaRoCr2015}} & {\cite{Woodham1980}}\\
\hline
\hline
 99.87\%  & 99.90 \%  & 96.31 \%  & 96.63 \% \\  
\hline
\end{tabular}
\vspace{2.0ex}
\centering
\caption{Our total speed improvement over other methods for all images of Table 1. It can be seen that our scheme can decrease consuming time dramatically as well as providing reliable results.}
\label{Table II}
\vspace{-5mm}
\end{table}

\begin{figure}[h]
\begin{center}
\begin{tabular}{c} 
\hspace{-5mm}
\includegraphics[width=0.38\textwidth]{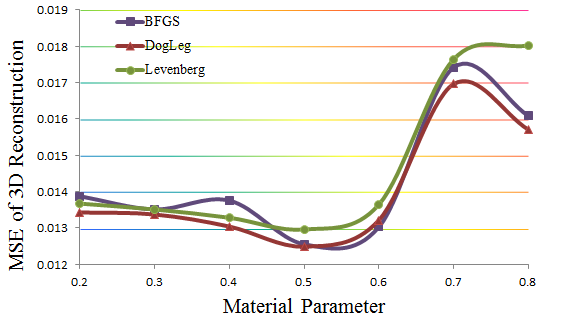}
\\ 
\vspace{3mm}(a) Oldman
\\
\hspace{-5mm}
\includegraphics[width=0.38\textwidth]{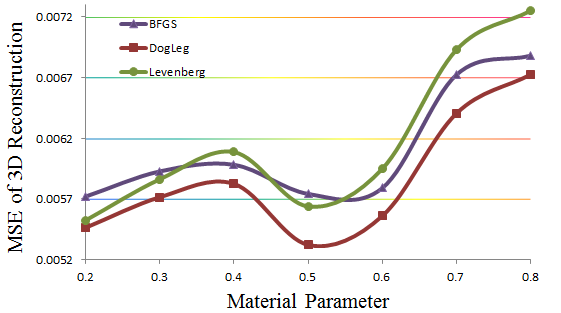}
\\ 
\vspace{3mm}(b) Militaryman 
\\
\hspace{-5mm}
\includegraphics[width=0.38\textwidth]{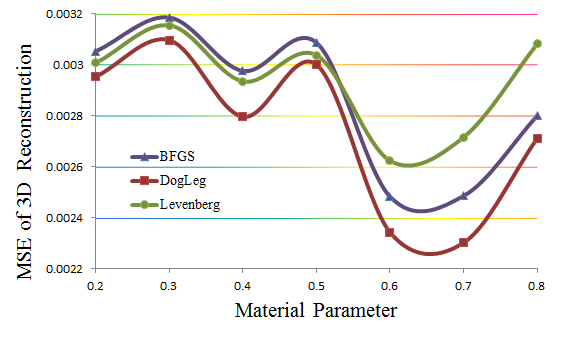}
\\ 
\vspace{3mm}(c) Carpenter 
\\
\includegraphics[width=0.38\textwidth]{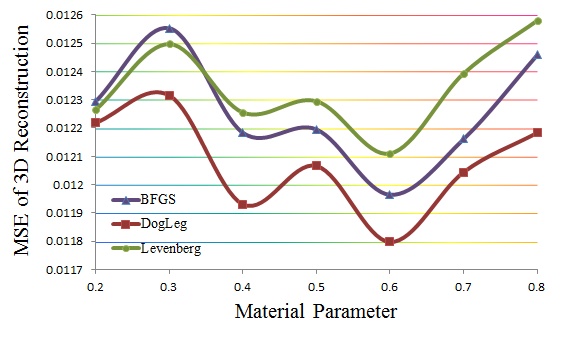}
\\ 
\vspace{3mm}(a) Woman 
\end{tabular}
\end{center}
\vspace{-3mm}
\caption{MSE of our 3D reconstructions with different optimization procedures for specular surfaces with varying material parameter ($k_s)$ for different input images.
\label{fig3}
}
\end{figure}
\subsection*{{Comparison with several PS methods:}}
\vspace{-2mm}\noindent As another evaluation shown in Table \ref{Table I}, we compared our method from different aspects on the set of images with the state-of-the-art approaches: constrained bivariate regression {\cite{IK14}}, differential ratios {\cite{MeccaRoCr2015}}, dimensionality reduction based on PCA {\cite{Barsky2003}} and also least squares-regression {\cite{Woodham1980}}. For a fair comparison, normal map are obtained by the finite-difference of depth map for the methods suggested in {\cite{{IK14}, MeccaRoCr2015}} (these schemes provide depth, but not normal vectors) and the depth map is obtained by our integrator (section 5.6.) for {\cite{{Barsky2003},{Woodham1980}}} (these techniques obtain normal vectors, but not depth map). All methods show worse performances than our approach. One reason is that high-frequency variations are mis-classified and leads to the unpredictable errors.
As can be seen in Table \ref{Table I}, the proposed strategy dramatically reduces error in computing normal vectors and depth map in comparison with other approaches thanks to our effective techniques of DGMC and RON and capability of handling specularities. The total accuracy of obtained normal field and depth map for all images applied in Table \ref{Table I} is shown in Figure \ref{diagrams}. For example, we could improve the accuracy in extracted normals by 93.15\% over {\cite{MeccaRoCr2015}} and 90.43 \% over {\cite{Woodham1980}} and provide higher accuracy in depth reconstruction by 73.29\% over {\cite{MeccaRoCr2015}} and 70.46 \% over {\cite{Woodham1980}}. Furthermore, the first row in Table \ref{Table I} demonstrates the efficiency of our proposed method. Our method is the first one with significant high speed providing all information of normal vectors, colorful albedo, roughness value and depth map. Our speed improvement over mentioned techniques is represented in Table 2.  
The running time of {\cite{MeccaRoCr2015,Woodham1980}} are comparable, yet their results appear both quantitatively and qualitatively different.    
The qualitative comparison of our approach with the mentioned methods is shown in Figure \ref{fig6}. The performance ranking of methods is consistent with qualitative results. Our results demonstrate the ability of the proposed scheme to provide a complete reconstruction without any deviation at specularities which outperforms other comparing approaches.  
\begin{table}[t!]
\begin{tabular}{ |l|l|l| }
\hline
\multicolumn{2}{ |c| }{MSE comparison for the applied Lightening models} \\
\hline
\hline
PLPS for Oldman      & \hspace{8.0ex}  0.007700 \\ \hline
DLPS for Oldman      & \hspace{8.0ex}  0.012408 \\ \hline
PLPS for Woman       & \hspace{8.0ex}  0.008439 \\ \hline
DLPS for Woman       & \hspace{8.0ex}  0.011801 \\ \hline
PLPS for Carpenter   & \hspace{8.0ex}  0.002580 \\ \hline
DLPS for Carpenter   & \hspace{8.0ex}  0.003085 \\ \hline
\end{tabular}
\vspace{2.0ex}
\centering 
\caption{MSE of our 3D shape reconstruction with perspective Cook-Torrance reflectance model in the presence of specularities using different lightening models.}
\label{Table II}
\vspace{-5mm}
\end{table}
\subsection{\bf\textit {Test of applicability on real-world images}}

Finally, we evaluated the capability of our approach on a set of real-world images 
shown in Figures \ref{fig7}, \ref{fig8}, \ref{fig9}, \ref{albedo} as a test of the practical applicability of our method. 
All the images used in these experiments are captured by the camera of a Samsung Galaxy S5 smartphone and using a flash. For image acquisition the photographed figurine is standing on a table covered
by a black tablecloth as shown in Figure \ref{fig1}. The light directions are measured by considering the camera as the origin of the coordinate system.
The recovered 3D shape for the first real-world image is presented in Figure \ref{fig7}. In this case, we created a complex scene composed of numerous small statues with different sizes as a challenging set. However, we could reconstruct the surface with correct geometry where relevant details and main features are recovered faithfully. Let us stress that our real-world images are taken without any laboratory equipment or controlled setups.
Nevertheless, the reconstruction result is in our opinion very reasonable. This test demonstrates the robustness of our method and its potential for real-world applications even using consumer-level equipment. Moreover, again for this real-world scene, we applied different PS-based lightnings models for a same light source of a flash. Improved quality of the 3D reconstruction with spatially dependent lights (PLPS) shown in Figure \ref{fig7} (in terms of the height field of depth map and details) can be an advantage offered by this model. More real-world experiments are presented in Figures \ref{fig8} and \ref{fig9}. As can be seen, our approach succeeds in producing accurate 3D reconstruction results even for high frequency details of tiny structure of Berlin souvenir statues. The colors of these images are extracted in Figure \ref{albedo} where the diversity of colors is recovered as well in our non-uniform colorful albedo reconstruction. 
Let us also point out that our results show the capability of reliable 3D reconstruction for 
specular materials by using a minimum number of input images (for all synthetic and real-world experiments, we used three input images). Thus we document here that our method works under reasonable practical conditions with inexpensive instruments so that it can be used for many potential applications.
\section{Conclusion}
We presented a real-time robust PS which is also benefiting from perspective Cook-Torrance reflectance model to explicitly handle specularities and remove the limitations of working with diffuse materials or orthographic projection. A {\bf{RON optimization technique}} based on {\bf{DGMC}} approach is introduced to obtain accurate information of normal vectors, depth, roughness and colorful albedo. Our proposed RON and DGMC techniques can be applied in any {\bf{highly non-linear optimization}} framework as a key point for providing multi-variable solutions and proper initialization. These innovations provide important steps towards a reliable PS. Furthermore, we equipped our model with the spatially dependent lightening offering more reasonable reconstructions. We have demonstrated the applicability of our method by applying minimum number of input images without any laboratory conditions or facilities. Furthermore, we stretch our approach to its full potential by extracting 3D reconstruction of micro-prints. Our method has the potential to be useful as the basis of future developments.
\clearpage
\begin{figure}[t!]
\centering
\begin{center}
\begin{tabular}{ c c } 
\includegraphics[width=0.22\textwidth]{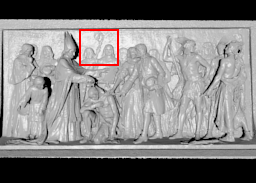}
&
\includegraphics[width=0.25\textwidth]{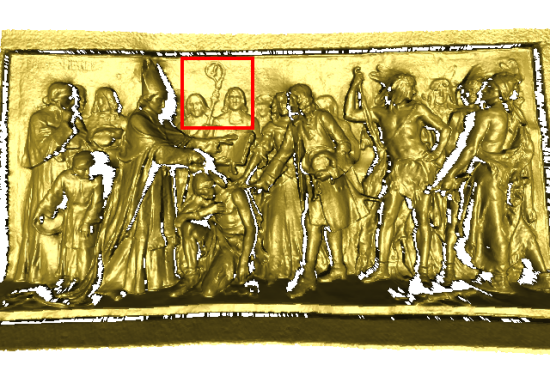}
\\
Coping
&
\vspace{-2mm}
\\
\includegraphics[width=0.2\textwidth]{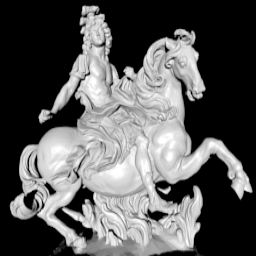}
&
\includegraphics[width=0.25\textwidth]{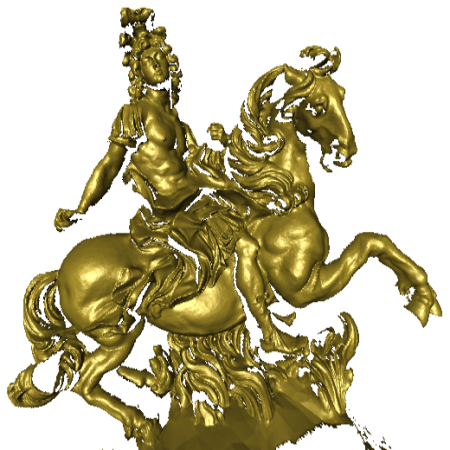}
\\
Horseman
&
\vspace{-2mm}
\\
\includegraphics[width=0.2\textwidth]{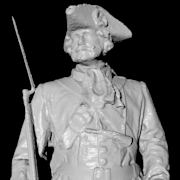}
&
\includegraphics[width=0.25\textwidth]{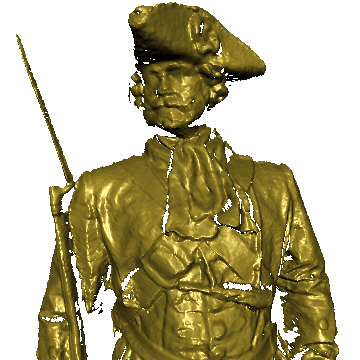}
\\
Soldier
&
\vspace{-2mm}
\\
\includegraphics[width=0.2\textwidth]{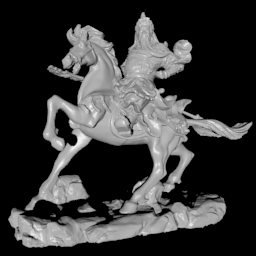}
&
\includegraphics[width=0.25\textwidth]{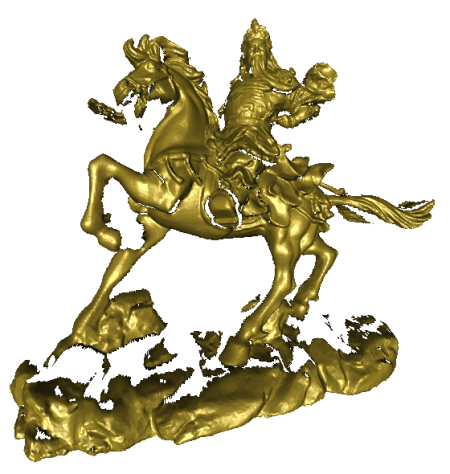}
\\
Fighter
&
\vspace{-2mm}
\\
\end{tabular}
\end{center}
\caption{Left: input images including specularity. Right: our 3D reconstruction results using perspective Cook-Torrance reflectance model. These results illustrate the capability of the proposed method for providing faithful reconstructions with high frequency details even for the fine details of a tiny surface (e.g. first input image).  
\label{fig4}
}
\vspace{-5mm}
\end{figure}
\begin{figure}[t!]
\begin{center}
\begin{tabular}{ccc} 
Carpenter
&
Oldman
&
Woman
\\
\includegraphics[width=0.13\textwidth]{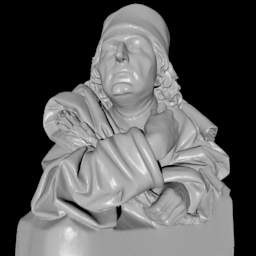}
&
\includegraphics[width=0.13\textwidth]{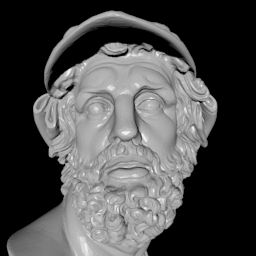}
&
\includegraphics[width=0.13\textwidth]{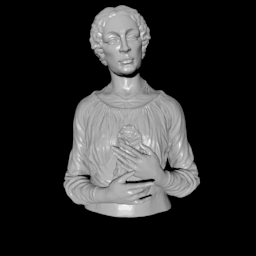}
\\
\includegraphics[width=0.14\textwidth]{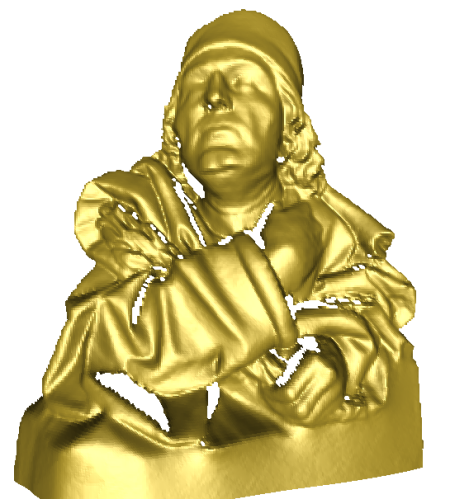}
&
\includegraphics[width=0.14\textwidth]{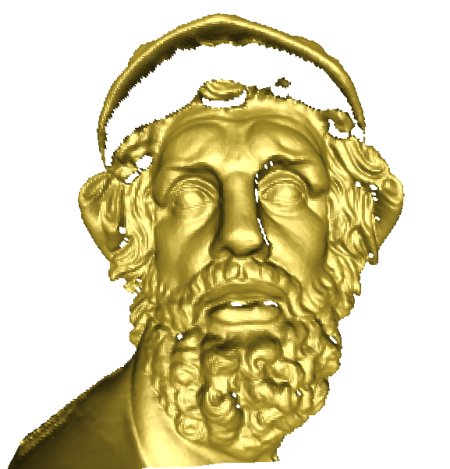}
&
\includegraphics[width=0.14\textwidth]{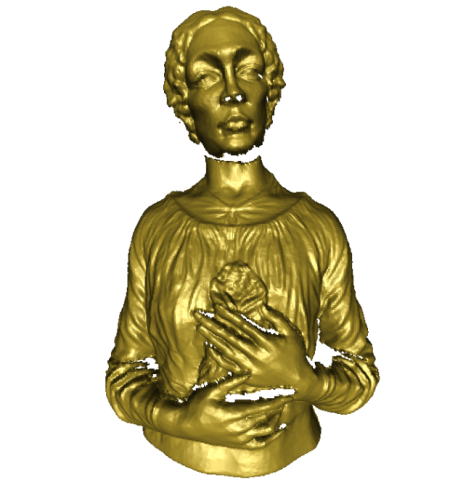}
\vspace{1mm}
\\
\includegraphics[width=0.13\textwidth]{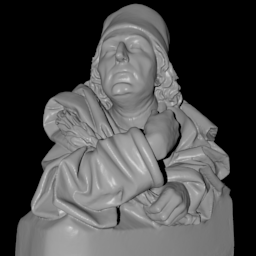}
&
\includegraphics[width=0.13\textwidth]{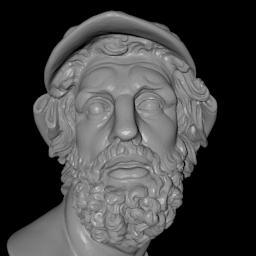}
&
\includegraphics[width=0.13\textwidth]{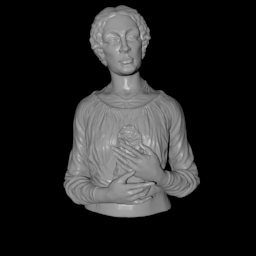}
\\
\includegraphics[width=0.14\textwidth]{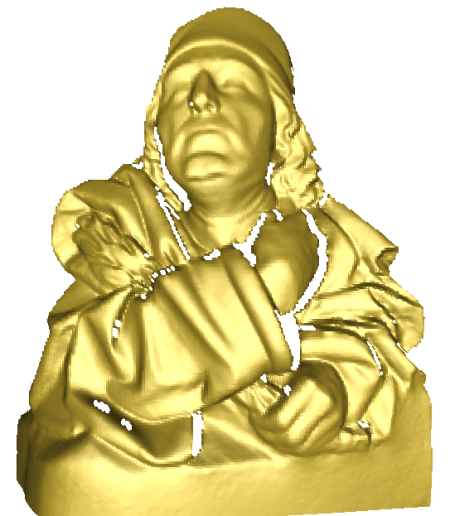}
&
\includegraphics[width=0.14\textwidth]{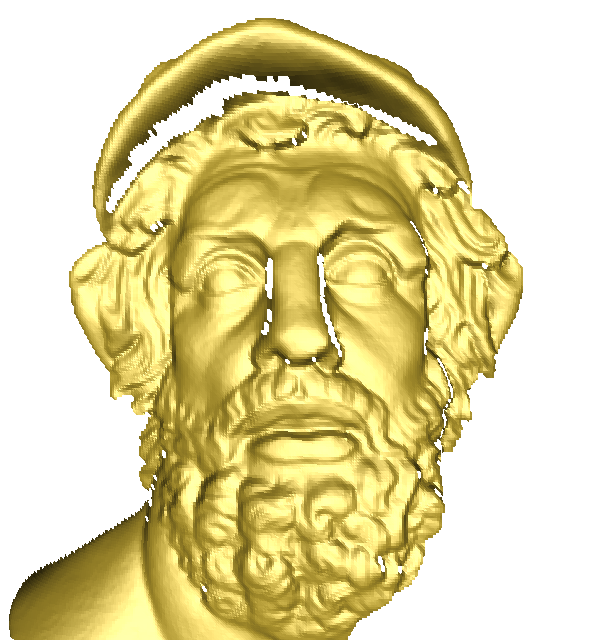}
&
\includegraphics[width=0.14\textwidth]{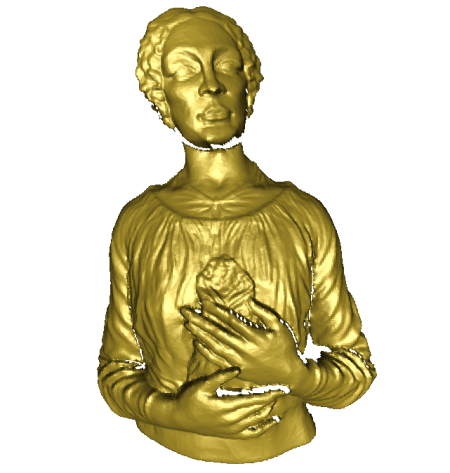}
\end{tabular}
\end{center}
\caption{ First row: input specular images with directional light. Second row: our 3D output using DLPS. Third row: input images with pointwise light. Last row: our 3D output with PLPS. Images produced by directional light are imposed by more specularity and shadows in comparison with images with pointwise light. These shadow parts lead to losing some reconstructed area. 
\label{fig5} 
}
\vspace{-8mm}
\end{figure}

\begin{figure*}[h!]
\centering
\begin{center}
\vspace{-3mm}
\begin{tabular}{ c c c c c c } 
\\
\includegraphics[width=0.14\textwidth]{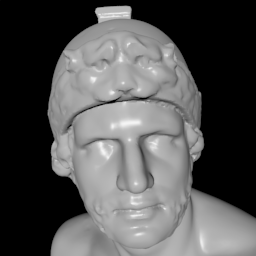}
&
\includegraphics[width=0.14\textwidth]{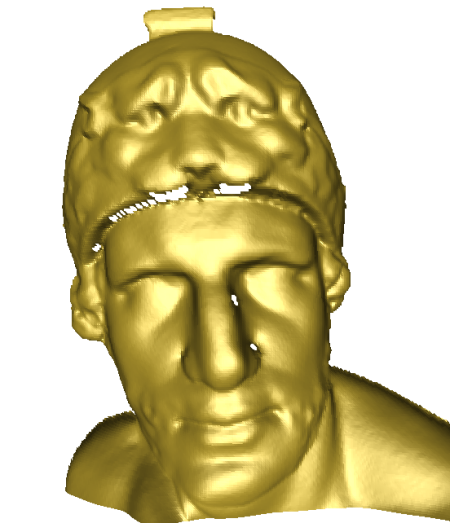}
&
\includegraphics[width=0.14\textwidth]{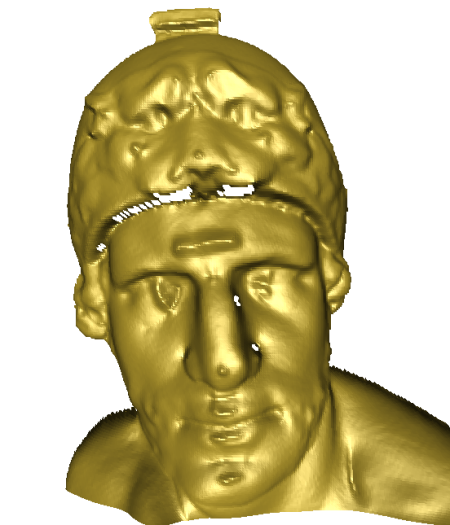}
&
\includegraphics[width=0.14\textwidth]{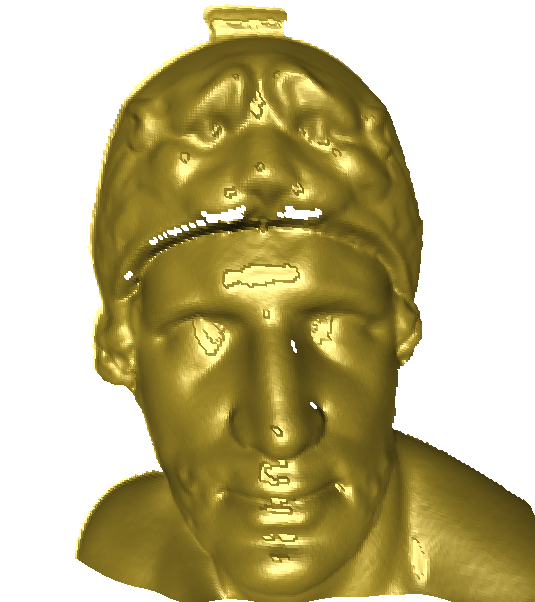}
&
\includegraphics[width=0.12\textwidth]{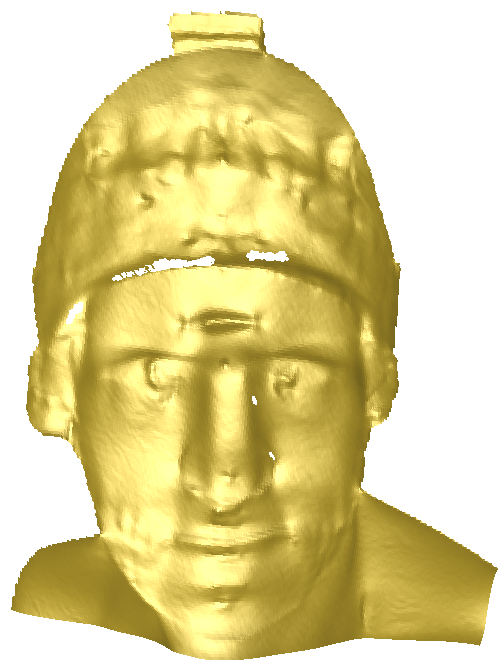}
&
\includegraphics[width=0.13\textwidth]{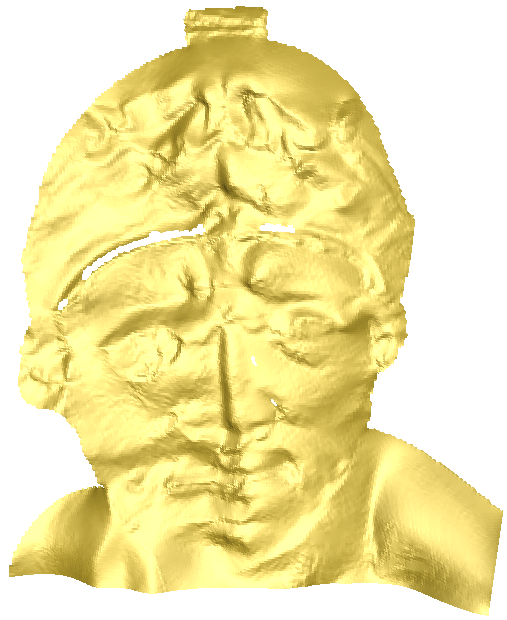}
\end{tabular}
\end{center}
\caption {{Comparison of several PS methods. From left: 1) input image (Militaryman) including high specularities, and the 3D results of 2) our reconstruction applying perspective Cook-Torrance reflectance model without any deviation or artifacts, 3) reconstruction of {\cite{Woodham1980}} that could not handle specularity areas, 4) reconstruction of \cite{MeccaRoCr2015} which shows deviations specially around the highly specular areas 5) {\cite{Barsky2003}} shows problem in producing complete 3D reconstruction in presence of spcularity and 5) {\cite{IK14}} which still includes distortions.}     
\label{fig6}
}
\end{figure*}
\begin{figure*}[h!]
\begin{center}
\vspace{-4mm}
\begin{tabular}{c c c c} 
\includegraphics[width=0.18\textwidth]{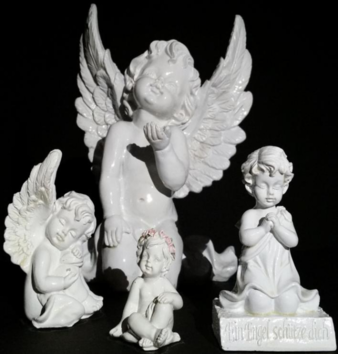}
&
\includegraphics[width=0.22\textwidth]{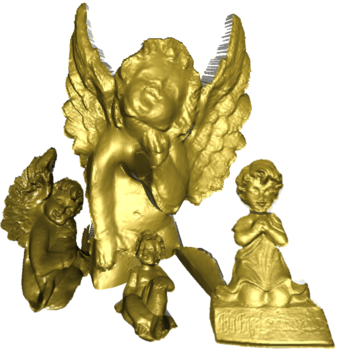}
&
\includegraphics[width=0.22\textwidth]{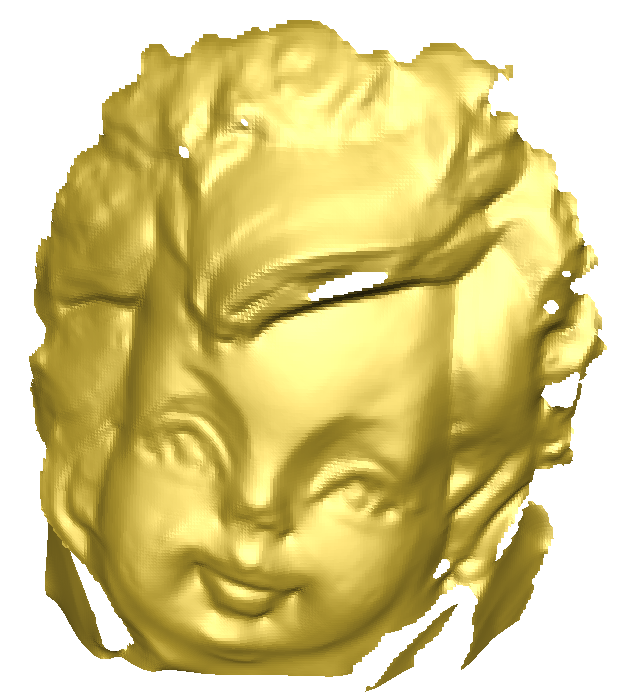}
&
\includegraphics[width=0.2\textwidth]{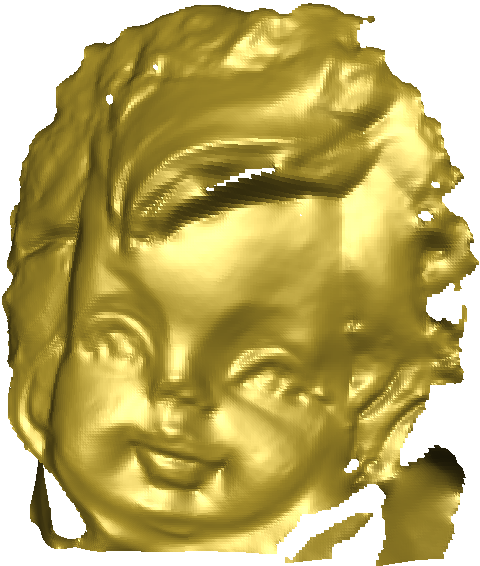}
\end{tabular}
\end{center}
\caption{From left: 1) input real-world image including several tiny statues captured without laboratory setting cf. Figure \ref{fig1}, 2) our 3D reconstruction with Cook-Torrance reflectance model, 3) our 3D reconstruction using DLPS for the face of the smallest statue in the scene and 4) our 3D reconstruction using PLPS for the same statue. 
\label{fig7}
}
\end{figure*}
\begin{figure*}[b!]
\centering
\begin{center}
\vspace{-3mm}
\begin{tabular}{c c c c} 
\includegraphics[width=0.135\textwidth]{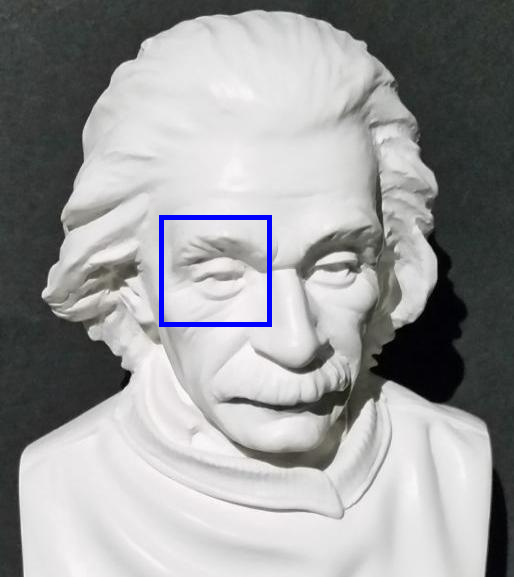}
&
\includegraphics[width=0.21\textwidth]{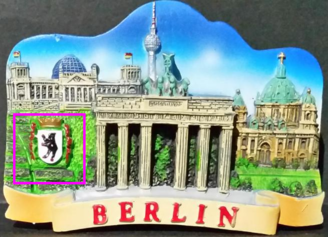}
&
\includegraphics[width=0.21\textwidth]{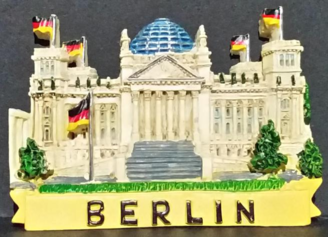}
&
\includegraphics[width=0.153\textwidth]{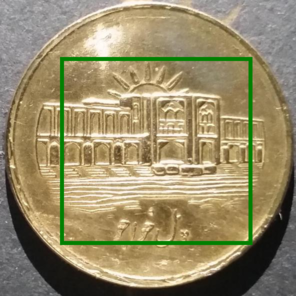}
\\
\includegraphics[width=0.18\textwidth]{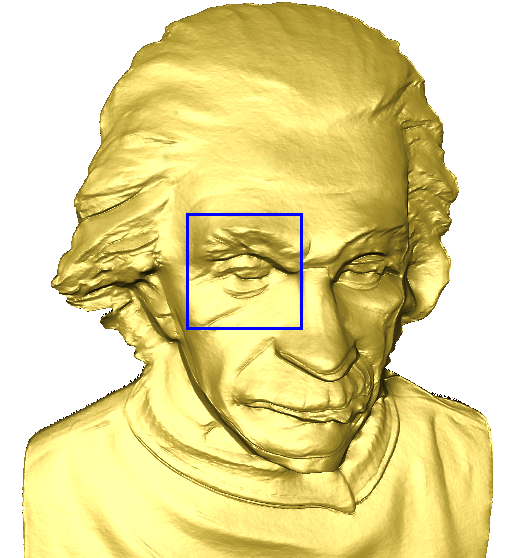}
&
\includegraphics[width=0.22\textwidth]{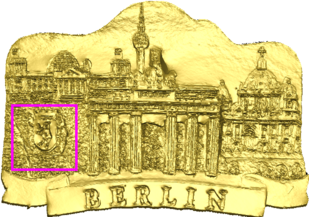}
&
\includegraphics[width=0.22\textwidth]{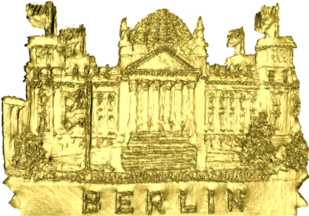}
&
\includegraphics[width=0.18\textwidth]{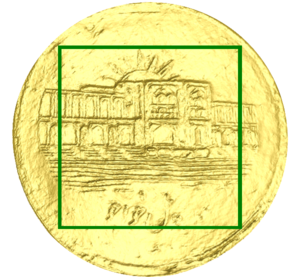}
\\
\includegraphics[width=0.18\textwidth]{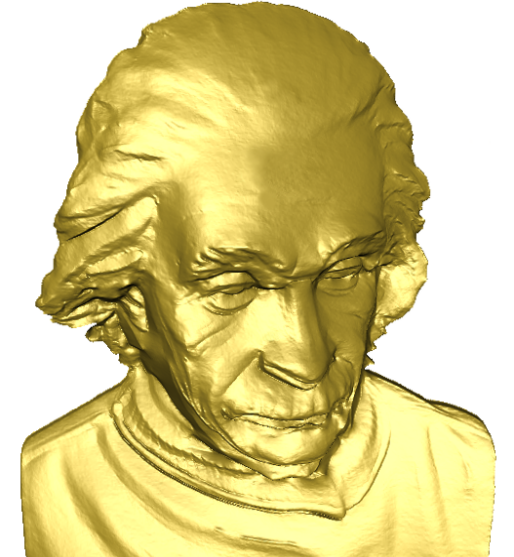}
&
\includegraphics[width=0.22\textwidth]{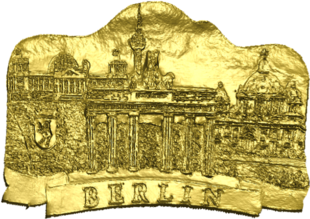}
&
\includegraphics[width=0.22\textwidth]{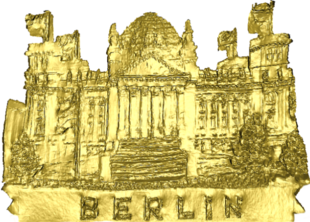}
&
\includegraphics[width=0.18\textwidth]{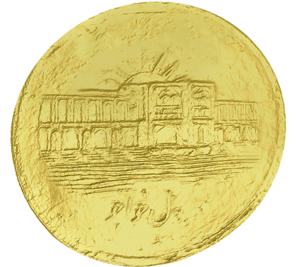}
\end{tabular}
\end{center}
\caption{More real-world experiments with specularity, first row) input images including Einstein statue, two Berlin souvenir statues with comprehensive structure and a metallic coin, second row) our 3D reconstruction using DLPS and last row) our 3D reconstruction using PLPS. High quality details and structures recovered by the proposed approach confirm offered advantages and agrees well with them.      
\label{fig8}
}
\end{figure*}
\begin{figure*}[h!]
\centering
\begin{center}
\vspace{-2mm}
\begin{tabular}{c c c c} 
\includegraphics[width=0.21\textwidth]{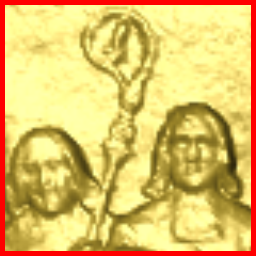}
&
\includegraphics[width=0.21\textwidth]{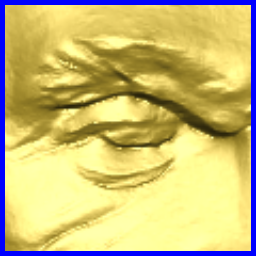}
&
\includegraphics[width=0.21\textwidth]{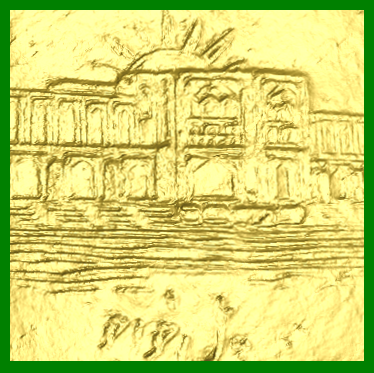}
&
\includegraphics[width=0.21\textwidth]{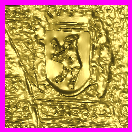}
\end{tabular}
\end{center}
\caption{Some qualitative evaluations for the proposed method: from left) close-up views of our reconstruction results corresponding to the red, blue, green and magenta rectangular areas in Coping as synthetic input and Einstein, Coin and first Berlin statue as the real scenes. It can be seen that our method achieves desirable reconstruction quality even for very fine details.           
\label{fig9}
}
\end{figure*}
\begin{figure*}[t!]
\centering
\begin{center}
\begin{tabular}{c c} 
\includegraphics[width=0.28\textwidth]{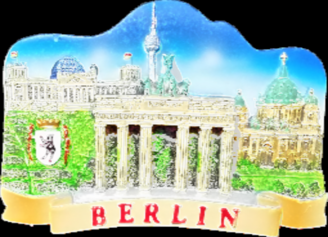}
&
\includegraphics[width=0.28\textwidth]{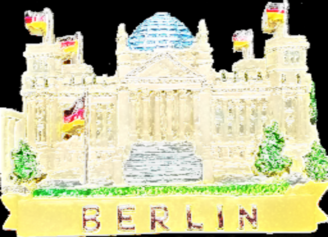}
\end{tabular}
\end{center}
\caption{Reconstructed albedo of two color input images shown in Figure 10. As can be seen, our RON approach can produce non-uniform albedo from images with a variety of colors. Since the albedo is not same across the entire real-world surfaces, our variant albedo extraction is more adjusted for real-world situations. It should be noticed that these statues are very tiny surfaces.       
\label{albedo}
}
\end{figure*}


\clearpage
\begin{algorithm*}
\LinesNotNumbered
\SetKwData{Left}{left}
\SetKwData{This}{this}
\SetKwData{Up}{up}
\SetKwFunction{Union}{Union}
\SetKwFunction{FindCompress}{FindCompress}
\SetKwInOut{Inputf}{Primary input}
\SetKwInOut{Inputs}{Secondary inputs}
\SetKwInOut{Output}{output}
\Inputf{$X^{0}$ that should be computed using our proposed "DGMC" technique explained in section 5.4. of the paper}
\Inputs{ ${\vartheta}$ = 0, k = 0}
\Output{Normal vectors for photometric stereo}

\BlankLine
\While{{$\Vert{J(X^{k})}\Vert>{\varepsilon}$, and $k<k_{max}$ and $\vartheta<\vartheta_{max}$ }}{
	{
		Compute {$d^{k}:=-({{B(X^k)}^{-1}})J(X^{k})$}\\				
	}
	{$[X^{k+1},d{\vartheta}]\leftarrow$ line search $(X^{k},d^{k})$} as applied in reference [33] of the paper\\{
	${\vartheta}$ $\leftarrow$ {${\vartheta}+d{\vartheta}$}
	\\
	$\Theta$ $\leftarrow$  {$X^{k+1}-X^{k}$}\\
	$Y$ $\leftarrow$  {$J({X}^{k+1})-J(X^{k})$}\\
	\If{$({{\Theta}})^{T}Y>{\sqrt{\varpi}}{\Vert{\Theta}\Vert}_{2}{\Vert{Y}\Vert}_{2}$}{
   $U$ $\leftarrow$ $B({X}^{k}){\Theta}$\\
   $B({X^{k+1}}):=B({X}^{k})+\frac{Y^{k}(Y^{k})^{T}}{({\Theta}^{k})^{T}Y^{k}}-\frac{U^{k}(U^{k})^{T}}{({\Theta}^{k})^{T}U^{k}}$
  }$k$ $\leftarrow$ $k+1$
	}
}{}
Where ${\varpi}$ is the computer accuracy and $\varepsilon$ = 1e -12 is considered.
\caption{Quasi-Newton with BFGS updating }\label{algo_disjdecomp}
\end{algorithm*}
\begin{algorithm*}
\LinesNotNumbered
\SetKwData{Left}{left}
\SetKwData{This}{this}
\SetKwData{Up}{up}
\SetKwFunction{Union}{Union}
\SetKwFunction{FindCompress}{FindCompress}
\SetKwInOut{Inputf}{Primary input}
\SetKwInOut{Inputs}{Secondary inputs}
\SetKwInOut{Output}{output}
\Inputf{$X^{0}$ that should be computed using our proposed "DGMC" technique explained in section 5.4. of the paper}
\Inputs{ ${\vartheta}$ = 2, $\lambda_k=\tau$  max$\{{a_{ii}\}}$, where $\{{a_{ii}\}}$ is the set of the diagonal elements of $A=J(X)J(X)^{T}$, $\Psi (X) = \frac{1}{2}{\Vert{F(X)}\Vert}^{2}_{2}$, k = 0}
\Output{Normal vectors for photometric stereo}

\BlankLine
\While{{$\Vert{J(X^{k})^{T}F(X^{k})}\Vert_{\infty}>{\varepsilon_{1}}$, and $k<k_{max}$ }}{
	{
		{$d^{k}:=-(J(X^{k})J(X^{k})^{T}+\lambda_{k}I)^{-1}J(X^{k})^TF(X^{k})$}\\					
	}

	\eIf{$\Vert d^{k}\Vert\ > \varepsilon_{2}(\Vert X^{k}\Vert+\varepsilon_{2}) $}{

	$X^{k+1}$ $\leftarrow$  {${X}^{k}+d^{k}$}
	\\
	$\aleph=\frac{\Psi(X^{k})-\Psi(X^{k+1})}{\frac{1}{2}(d^{k})^{T}\big(\lambda_k(d^{k})-J(X^{k})^{T}F(X^{k})\big)}$\\
	\eIf{$\aleph > 0 $}{
   $\lambda_{k+1}=\lambda_k$ max $\big\{ \frac{1}{3}, 1-(2\aleph -1)^{3}\big\}$, $\vartheta=2$}
   {
   $\lambda_{k+1}=\lambda_k\vartheta, \;\vartheta=2\vartheta$\\
  }	
}{Exit}
$k$ $\leftarrow$ $k+1$
}
{{}}
Where $\tau$ = 9e-2 and $\varepsilon_1 = \varepsilon_2 =$ 1e-15 are considered.
\caption{Levenberg-Marquardt}\label{algo_disjdecomp}
\end{algorithm*}
\begin{algorithm*}
\LinesNotNumbered
\SetKwData{Left}{left}
\SetKwData{This}{this}
\SetKwData{Up}{up}
\SetKwFunction{Union}{Union}
\SetKwFunction{FindCompress}{FindCompress}
\SetKwInOut{Inputf}{Primary input}
\SetKwInOut{Inputs}{Secondary inputs}
\SetKwInOut{Output}{output}
\Inputf{$X^{0}$ that should be computed using our proposed "DGMC" technique explained in section 5.4. of the paper}
\Inputs{ ${\Delta}=\Delta_{0}$, $\Psi (X) = \frac{1}{2}{\Vert{F(X)}\Vert}^{2}_{2}$, k = 0}
\Output{Normal vectors for photometric stereo}

\BlankLine
\While{{$\Vert{F(X^{k})}\Vert_{\infty}>{\varepsilon_{1}}$ and {$\Vert{J(X^{k})^{T}F(X^{k})}\Vert_{\infty}>{\varepsilon_{2}}$} and ($k<k_{max}$)}}{
	{
	\vspace{2mm}
		{$\alpha=\frac{{\Vert{J(X^{k})^{T}F(X^{k})}\Vert}^{2}}{{\Vert{J(X^{k})J(X^{k})^{T}F(X^{k})}\Vert}^{2}}$}\\
\vspace{2mm}		$h_{sd}=-\alpha{J(X^{k})^{T}F(X^{k})}$\\	
\vspace{2mm}		$h_{gn}=-{J(X^{k})^{-1}F(X^{k})}$\\
		
	     \[\hspace{-90mm}h_{dl}=\left\{ \begin{array}{ll}
	     \vspace{2mm}
         h_{gn} & \mbox{ $\Vert{h_{gn}}\Vert \leq \Delta$}\\
         \vspace{2mm}
         \frac{\Delta h_{sd}}{\Vert h_{sd}\Vert} & \mbox{$\Vert \alpha{h_{sd}}\Vert \geq \Delta$}\\
         \alpha h_{sd}+ \beta (h_{gn}-\alpha h_{sd})&  \quad \;\;\;\mbox{o.w.}
         
         \end{array} \right. \] 
								
	}

	\eIf{$\Vert h_{dl}\Vert > \varepsilon_{3}(\Vert X^{k}\Vert+\varepsilon_{3}) $}{

	$X^{k+1}$ $\leftarrow$  {${X}^{k}+ h_{dl}$}
	\\
	$\aleph=\frac{\Psi(X^{k})-\Psi(X^{k+1})}{L(0)-L(h_{dl})}$\\
  \uIf{$\aleph > 0.75 $}{
   $\Delta=$ max $\big\{ \Delta, {3}{\Vert h_{dl}\Vert}\big\}$
  }
  \ElseIf{$\aleph < 0.25 $}{
   $\Delta=\frac{\Delta}{2}$\\   
  }
  \If {$\Delta \leq \varepsilon_{3}(\Vert X^{k}\Vert+\varepsilon_{3})$}{Exit}
}{Exit}  
}{{}}

\vspace{5mm}
where
\\
$L(0)- L(h_{dl})= \left\{ \begin{array}{ll}
       \vspace{2mm}
      \Psi(X)& \quad\quad\mbox{${\hspace{-3ex}}if \hspace{0.9ex}{h_{dl}}={h_{gn}}$}\\
       \vspace{2mm}
         \frac{\Delta  \bigg(2\Vert \alpha {J(X^{k})^{T}F(X^{k})\Vert}-\Delta \bigg)}{2\alpha} & \mbox{$if \hspace{0.9ex} {h_{dl}=\frac{-\Delta J(X^{k})^{T}F(X^{k})}{\Vert J(X^{k})^{T}F(X^{k}) \Vert}}$}\\
        \frac{1}{2}\alpha {(1- \beta)}^{2} {\Vert J(X^{k})^{T}F(X^{k}) \Vert}^{2}+\beta(2-\beta) F(X)& \quad \quad\quad\mbox{o.w.}
         
         \end{array} \right. $
         \vspace{5mm}
         \\and for choosing $\beta$, the strategy proposed by reference [33] is applied.
         \\In addition, $\Delta_0$ = 5e-1 and $\varepsilon_1$ = $\varepsilon_2$ = $\varepsilon_3$ = 1e-12.
         
\caption{Powell's Dog Leg}\label{algo_disjdecomp}
\end{algorithm*}

\clearpage
\bibliographystyle{CVM}
{\normalsize \bibliography{ref}}

\end{document}